\documentclass[letterpaper]{article} 
\usepackage{aaai2026}  
\usepackage{times}  
\usepackage{helvet}  
\usepackage{courier}  
\usepackage[hyphens]{url}  
\usepackage{graphicx} 
\usepackage{natbib}  
\usepackage{caption} 
\usepackage{subcaption}
\usepackage{booktabs}
\usepackage{multirow}
\usepackage{amsmath} 
\usepackage{float}
\usepackage{amsfonts}
\usepackage{array}
\usepackage{makecell}
\usepackage{algorithm}
\usepackage{algorithmic}

\usepackage{newfloat}
\usepackage{listings}
\DeclareCaptionStyle{ruled}{labelfont=normalfont,labelsep=colon,strut=off} 
\floatstyle{ruled}
\newfloat{listing}{tb}{lst}{}
\floatname{listing}{Listing}
\title{IndexTTS2: A Breakthrough in Emotionally Expressive and Duration-Controlled Auto-Regressive Zero-Shot Text-to-Speech}
\author{
    Siyi Zhou\textsuperscript{\rm 1},
    Yiquan Zhou\textsuperscript{\rm 1},
    Yi He\textsuperscript{\rm 1},
    Xun Zhou\textsuperscript{\rm 1},
    Jinchao Wang\textsuperscript{\rm 1},
    Wei Deng\textsuperscript{\rm 1},
    Jingchen Shu\textsuperscript{\rm 1}
}

\affiliations{
    \textsuperscript{\rm 1}Artificial Intelligence Platform Department, bilibili, China\\
    zhousiyi02@bilibili.com, zhouyiquan01@bilibili.com, heyi05@bilibili.com, zhouxun@bilibili.com, wangjinchao@bilibili.com, xuanwu@bilibili.com, shujingchen@bilibili.com, 
}

\usepackage{bibentry}

\begin{document}

\maketitle

\begin{abstract}
Existing autoregressive large-scale text-to-speech (TTS) models have advantages in speech naturalness, but their token-by-token generation mechanism makes it difficult to precisely control the duration of synthesized speech. This becomes a significant limitation in applications requiring strict audio-visual synchronization, such as video dubbing. This paper introduces IndexTTS2, which proposes a novel, general, and autoregressive model-friendly method for speech duration control. The method supports two generation modes: one explicitly specifies the number of generated tokens to precisely control speech duration; the other freely generates speech in an autoregressive manner without specifying the number of tokens, while faithfully reproducing the prosodic features of the input prompt. Furthermore, IndexTTS2 achieves disentanglement between emotional expression and speaker identity, enabling independent control over timbre and emotion. In the zero-shot setting, the model can accurately reconstruct the target timbre (from the timbre prompt) while perfectly reproducing the specified emotional tone (from the style prompt). To enhance speech clarity in highly emotional expressions, we incorporate GPT latent representations and design a novel three-stage training paradigm to improve the stability of the generated speech. Additionally, to lower the barrier for emotional control, we designed a soft instruction mechanism based on text descriptions by fine-tuning Qwen3, effectively guiding the generation of speech with the desired emotional orientation. Finally, experimental results on multiple datasets show that IndexTTS2 outperforms state-of-the-art zero-shot TTS models in terms of word error rate, speaker similarity, and emotional fidelity. Audio samples are available at:

\textbf{https://index-tts.github.io/index-tts2.github.io/}.
\end{abstract}


\section{Introduction}
\begin{figure*}[tp]
    \centering
    \includegraphics[width=\textwidth]{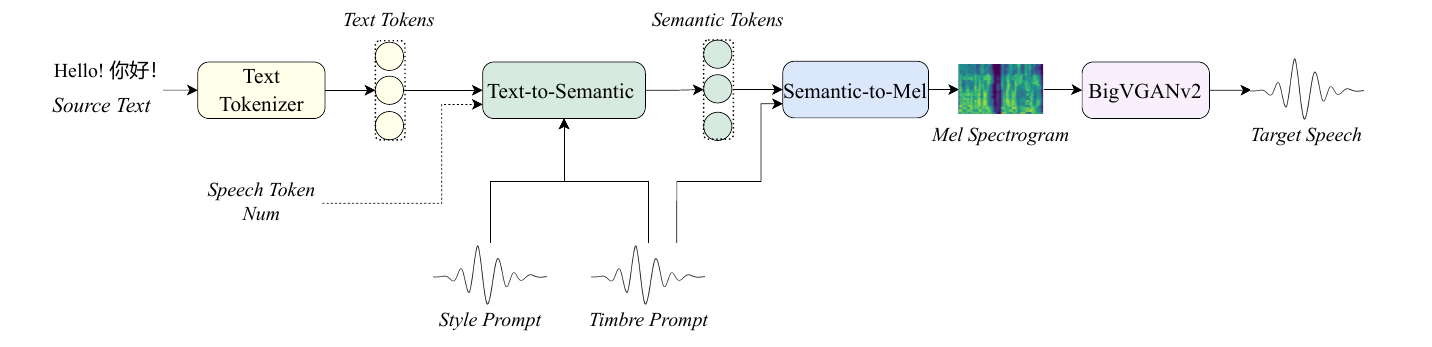}
    \caption{The overview of IndexTTS2.}
    \label{fig:overview_figure}
\end{figure*}

Recent advances in vector quantization \cite{vq,fsq}, Transformer architectures \cite{transformer,llama}, and large-scale data have enabled zero-shot TTS models to synthesize speech with timbre, prosody, and emotion from minimal audio prompts \cite{shen2023naturalspeech,casanova2024xtts,du2024cosyvoice2}. These models outperform traditional systems \cite{fastspeech2,kim2020glow} in naturalness and flexibility, enabling applications like AI dubbing \cite{cong2025emodubber}. Current TTS models are categorized into autoregressive (AR) \cite{sahipjohn2024dubwise,li2025flespeech,kim2023sc,du2024cosyvoice,zhou2024voxinstruct,chen2024takin,wang2025spark,guo2024fireredtts} and non-autoregressive (NAR) \cite{chen2025drawspeech,lee2024ditto,shen2023naturalspeech,yang2024simplespeech,wang2024maskgct,le2023voicebox,e2tts} systems. AR-based zero-shot TTS models like XTTS \cite{casanova2024xtts}, Cosyvoice \cite{du2024cosyvoice,du2024cosyvoice2}, and SparkTTS \cite{wang2025spark} show significant performance in terms of naturalness and expressiveness owing to their random sampling strategy and token-by-token generation. NAR-based models such as MaskGCT \cite{wang2024maskgct} and F5-TTS \cite{chen2024f5} enable fast inference via parallel decoding and support flexible parameter control (e.g., duration) through human intervention or model autonomy. However, AR models face challenges in duration control due to their sequential generation nature, limiting their applicability in time-sensitive scenarios like automated dubbing. Additionally, while TTS models excel in timbre reproduction, their emotional expression remains limited by scarce training data. Existing methods for emotional expression include emotion labels in training data \cite{ZhouSRSL23,PAVITS}, mapping natural language descriptions with emotion audio via CLAP \cite{clap,clip}, instruction fine-tuning \cite{du2024cosyvoice2}, and reference to emotional audio \cite{zhou2024voxinstruct}, but these approaches lack robustness in affective range and control precision.  

We introduce IndexTTS2 (Figure \ref{fig:overview_figure}), a novel zero-shot speech generation model that addresses both fixed-duration speech generation and natural-duration speech synthesis while enhancing emotional expressiveness. The model comprises three core modules: the Text-to-Semantic (T2S) module, the Semantic-to-Mel (S2M) module, and the Vocoder. The T2S module employs an autoregressive transformer framework to generate semantic tokens from text, timbre/style prompts, and an optional speech token count. Under specified token count constraints, a duration encoding mechanism ensures fixed-length token sequences with preserved semantic integrity. For emotional modeling, the T2S module extracts emotional features from style prompts and uses a Gradient Reversal Layer (GRL) \cite{ganin2016domain_grl,ju2024naturalspeech3zeroshotspeech} to eliminate emotion-irrelevant information during training. A multi-stage training strategy is adopted to overcome the lack of high-quality emotional data and enhance expressive capabilities. To enable natural language emotion control in speech synthesis, we further design a Text-to-Emotion (T2E) module, distilling Deepseek-r1's \cite{guo2025deepseekr1} emotion distribution prediction ability into Qwen-3-1.7b \cite{yang2025qwen3technicalreport} via Low-Rank Adaptation (LoRA) \cite{lorasurvey,devalal2018lora,bor2016lora}, and combine these probabilities with precomputed emotion embeddings to condition the T2S output. The S2M module generates mel-spectrograms via a non-autoregressive architecture, incorporating GPT latent representations to stabilize speech clarity during intense emotional expressions. The Vocoder module utilizes BigVGANv2 \cite{lee2022bigvgan} to convert mel-spectrograms into audio waveforms. 

The key contributions of this work are:  
\begin{itemize}  
\item[$\bullet$] We propose a duration adaptation scheme for autoregressive TTS models. IndexTTS2 is the first autoregressive zero-shot TTS model to combine precise duration control with natural duration generation, and the method is scalable for any autoregressive large-scale TTS model.  
\item[$\bullet$] The emotional and speaker-related features are decoupled from the prompts, and a feature fusion strategy is designed to maintain semantic fluency and pronunciation clarity during emotionally rich expressions. Furthermore, a tool was developed for emotion control, utilising natural language descriptions for the benefit of users.  
\item[$\bullet$] To address the lack of highly expressive speech data, we propose an effective training strategy, significantly enhancing the emotional expressiveness of zeroshot TTS to State-of-the-Art (SOTA) level.  
\item[$\bullet$] We will publicly release the code and pre-trained weights to facilitate future research and practical applications.  
\end{itemize}

\begin{figure*}[htp]
    \centering
    \includegraphics[width=0.8\textwidth]{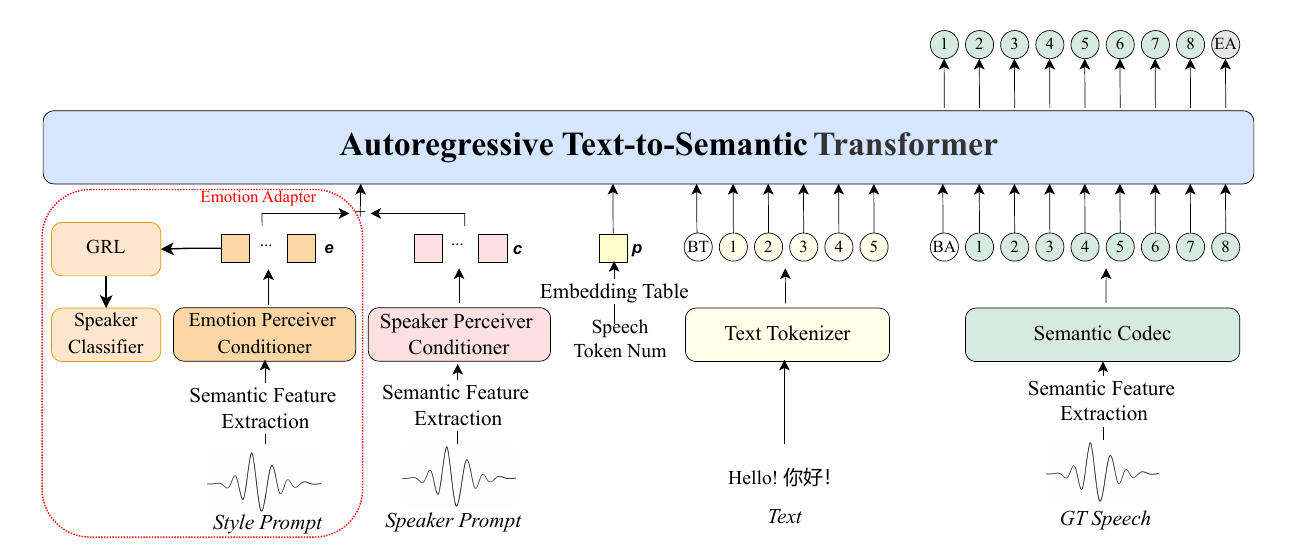}
    \caption{Autoregressive Text-to-Semantic Module. When speech token num is specified, precise control of the number of synthesized semantic tokens is performed. The emotion adapter (red dashed lines) is employed to extract emotional features from the style prompt, which are then used as input to the Text-to-Semantic process for the reconstruction of emotions.}
    \label{fig:gpt_figure}
\end{figure*}
\section{Related Work}
\subsubsection{Precise Duration Control for Large-Scale TTS.} Current zero-shot large-scale TTS models adopt autoregressive or non-autoregressive paradigms, with non-autoregressive approaches excelling in duration control via duration predictors based on diffusion, transformers \cite{lee2024ditto}, flows \cite{kim2023sc}, or language models \cite{yang2024simplespeech}. Methods like MaskGCT \cite{wang2024maskgct} use flow modeling for phoneme-level duration predictors based on diffusion, transformers \cite{lee2024ditto}, flows \cite{kim2023sc}, or language models \cite{yang2024simplespeech}. Methods like MaskGCT \cite{wang2024maskgct} use flow modeling for phoneme-level durations, while F5-TTS \cite{chen2024f5} estimates durations via text-speech length ratios. Autoregressive models (e.g., VoxInstruct \cite{zhou2024voxinstruct}, Takin \cite{chen2024takin}) rely on natural language instructions but face precision limitations. Techniques like CosyVoice \cite{du2024cosyvoice}, Spark-TTS \cite{wang2025spark}, DubWise \cite{sahipjohn2024dubwise}, and FleSpeech \cite{li2025flespeech} address token generation control through specialized cues, attribute labels, cross-modal fusion, or multimodal embeddings. This work introduces an enhanced autoregressive TTS model with precise token number control to overcome these challenges.

\subsubsection{Emotionally Controllable Large-Scale TTS.} Emotion control in large-scale TTS leverages natural language descriptions (e.g., Controlspeech \cite{ji2024controlspeech}), with CosyVoice \cite{du2024cosyvoice} using preset instructions and EmoSphere++ \cite{cho2025emosphere++} generating interpretable style embeddings. StyleTTS 2 \cite{li2023styletts} employs diffusion-based style vectors, SC VALL-E \cite{kim2023sc} integrates style networks, and Vevo \cite{zhang2025vevo} uses content-style token systems. Multimodal approaches like FleSpeech \cite{li2025flespeech} embed textual, audio, and visual cues into unified representations for precise regulation. This work enhances emotional expressiveness via additional emotion features, enabling flexible control through natural language or reference audio inputs.
\section{Proposed Method}
We propose IndexTTS2, a cascaded autoregressive zero-shot TTS system comprising three modules: the Text-to-Semantic (T2S) module, Semantic-to-Mel (S2M) module, and BigVGANv2 vocoder, each trained separately with tailored strategies to enhance emotional expressiveness. The T2S module generates semantic tokens from target text, style/timbre prompts, and an optional speech token count, while the S2M module predicts mel-spectrograms using these tokens and the timbre prompt. The BigVGANv2 vocoder then converts the mel-spectrograms into speech waveforms. To enable natural language-based emotional control, we introduce a Text-to-Emotion (T2E) module that produces an emotion vector from input text, which is integrated into the T2S module via a dedicated emotion vector interface. This design facilitates flexible, high-quality emotional speech synthesis through explicit natural language instructions or reference audio inputs.

\subsection{Autoregressive Text-to-Semantic Module (T2S)}

We formulate T2S as an autoregressive semantic token prediction task. As shown in Figure \ref{fig:gpt_figure}, the input sequence is constructed as $[c, p,  e_{\langle BT\rangle}, E_{text},  e_{\langle BA\rangle}, E_{sem}]$, where $c$ denotes speaker attributes, $p$ controls duration, $E_{text}$ represents text embeddings, and $E_{sem}$ denotes the embeddings of semantic tokens extracted from ground-truth speech via a semantic codec. $e_{\langle BT\rangle}$ and $e_{\langle BA\rangle}$ function as dedicated boundary tokens, serving to demarcate the extents of the text sequence and the semantic sequence, respectively. Our architecture resembles IndexTTS \cite{deng2025indextts} with key innovations in duration and emotion control.

\subsubsection{Duration Control:}
Duration regulation is achieved through a dedicated embedding $p$ computed from the target semantic token length $T$, where $p = W_{\text{num}} h(T)$. Here, $W_{\text{num}} \in \mathbb{R}^{L_{\text{speech}} \times D}$ represents an embedding table with $L_{\text{speech}}$ denoting the maximum semantic sequence length and $D$ being the embedding dimensionality. The function $h(T)$ returns a one-hot vector corresponding to $T$ \cite{onehot}. In particular, we implemented a special trick. We set the constraint $W_{\text{sem}} = W_{\text{num}}$ is imposed between $W_{\text{num}}$ and the semantic positional embedding table $W_{\text{sem}}$. This equation enables the autoregressive system to precisely align positional information with target duration information during generation, thereby accurately producing sequences of the desired length.

\subsubsection{Emotional Control:}
Emotion synthesis integrates an emotion embedding $e$ into the conditioning feature via the input sequence $[c + e,  p,  e_{\langle BT\rangle}, E_{\text{text}},  e_{\langle BA\rangle}, E_{\text{sem}}]$, where $e$ is extracted from style prompts using a Conformer-based emotion perceiver conditioner. To effectively capture the representation of emotional rhythm, we employ the following design: the speaker feature $c$ is derived from a pre-trained speaker perceiver conditioner extractor and primarily encodes timbral characteristics. To minimize the content overlap between $e$ and $c$ while enhancing feature disentanglement, we employ a GRL during training. This adversarial mechanism forces $e$ to exclusively capture emotional and rhythmic attributes, remaining invariant to speaker-specific timbre characteristics, thereby enabling more precise and robust control over global emotional prosody generation.

\subsubsection{Training and Inference:}
Our training data is organized by speaker, with each speaker having at least two utterances.
For prompt and target partitioning, we divide different utterances from the same speaker into prompts and training targets. 
To enhance data diversity, we apply random speed perturbation to both real speech and prompts using scaling coefficients $r_1$ and $r_2$.

We employ a dedicated three-stage training strategy for the T2S module:

\textbf{Stage 1:} The model is trained on the full dataset using the input sequence 
$
[c, p,  e_{\langle BT\rangle}, E_{text},  e_{\langle BA\rangle}, E_{sem}]
$,
where $c$ is the speaker embedding and $p$ is the duration embedding. 
To support both duration-controlled and free-form generation, $p$ is randomly set to zero with a probability of 30\%. 
This stage establishes the model's foundational capabilities.

\textbf{Stage 2:} We refine the emotion control module using the modified input sequence 
$
[c+e,p, e_{\langle BT\rangle}, E_{\text{text}}, e_{\langle BA\rangle}, E_{\text{sem}}]
$,
where $e$ denotes the emotion embedding. 
In this stage, the speaker perceiver conditioner (producing $c$) is frozen, while the emotion perceiver conditioner remains trainable. 
To disentangle speaker identity from emotional expression, a GRL and a speaker classifier are applied. 
Training is conducted on a curated subset of 135 hours of high-quality emotional speech. 
The joint loss function is defined as
\begin{equation}
L_{\text{AR}} = -\frac{1}{T+1}\sum_{t=0}^{T}\log q(y_t) - \alpha \log q(e),
\end{equation}
where $y_T$ represents the end-of-sequence token $⟨EA⟩$, $q(y_t)$ denotes the posterior probability of semantic tokens, $q(e)$ denotes the posterior probability that $e$ originates from the target speaker and $\alpha$ is the loss coefficient.

\textbf{Stage 3:} To improve robustness, we freeze all feature conditioners and perform fine-tuning on the full dataset.

During inference, duration control is achieved by setting $p = W_{\text{num}}h(T)$, while free-form generation is enabled by using $p = \mathbf{0}$. 
Emotional prosody can be directly manipulated by providing a desired emotion vector $e$ as input.
\begin{figure}[htp]
    \centering
    \includegraphics[width=0.48\textwidth]{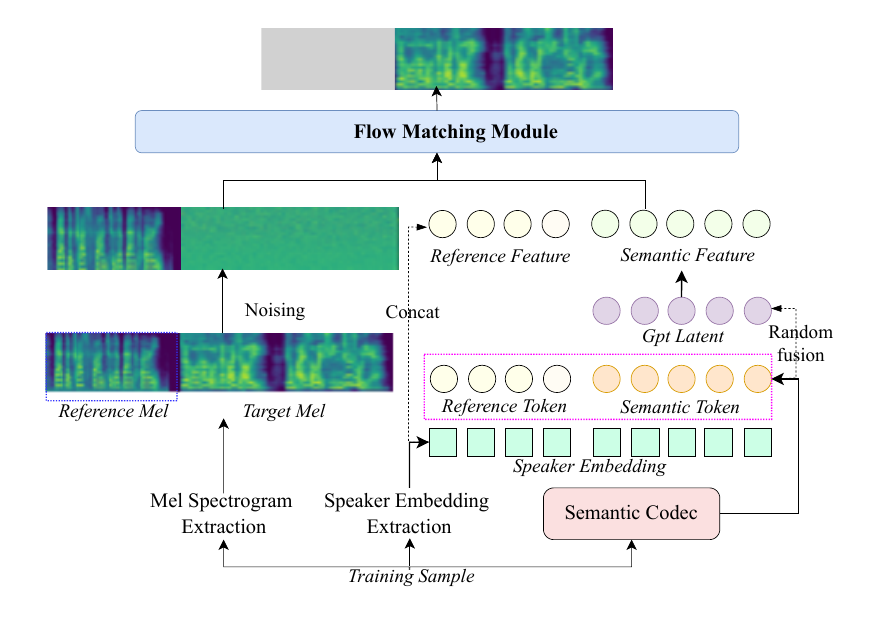}
    \caption{Semantic-to-Mel module based on flow matching.}
    \label{fig:seedvc_figure}
\end{figure}

\subsection{Semantic-to-Mel Module (S2M)}

As shown in Figure \ref{fig:seedvc_figure}, the S2M module employs a non-regressive generation framework based on flow matching \cite{flowmatching,seedvc,peebles2023scalable}. The model synthesizes target mel-spectrograms by combining prompt mel-spectrograms, speaker embeddings, and semantic features. To address pronunciation issues in emotional speech generation, we introduce GPT latent enhancement.

\subsubsection{GPT Latent Enhancement:}
Conditional flow matching learns an Ordinary Differential Equation model~\cite{ode} that maps samples from a noise distribution to target mel-spectrograms, conditioned on timbre reference audio and semantic codes generated by the T2S module. 

To mitigate speech slurring in speech synthesis, especially when synthesizing emotional speech, we introduce a novel approach leveraging latent features from the GPT model, denoted as $H_{\text{GPT}}$, which are extracted from the output of the final transformer layer in the T2S module. Given that the T2S module is trained to convert text into rich semantic representations using a large-scale dataset, we hypothesize that $H_{\text{GPT}}$ encodes substantial textual and contextual information. To exploit this, we fuse $H_{\text{GPT}}$ with the semantic features via vector addition, forming an enhanced, context-enriched representation. This fused feature is then used as input to the S2M training process.
Ablation studies validate that this integration effectively reduces the word error rate in highly expressive speech synthesis.


\subsubsection{Training and Inference:}
The S2M module is trained in a single stage. During training, each input sentence is randomly split into a prompt segment and a target segment. The mel-spectrograms corresponding to the target segment are fully noised to form the source inputs for the diffusion process. Semantic tokens generated by the T2S module are denoted as $Q_{sem}$. 
To improve pronunciation robustness, a Multi-Layer Perceptron (MLP)~\cite{rosenblatt1958perceptron,rumelhart1986learning} is employed to randomly fuse the GPT hidden states $H_{\text{GPT}}$ and the semantic tokens $Q_{sem}$ with 50\% probability, forming the final semantic representation $Q_{fin}$. Speaker embeddings, extracted via a perceiver-based conditioner, are concatenated with $Q_{fin}$ to ensure timbre consistency. The model is optimized using L1 loss~\cite{l1loss1} between the predicted ($y_{pred}$) and target ($y_{tar}$) mel-spectrograms:
\begin{equation}
    \mathcal{L}_{L1} = \frac{1}{F \cdot D} \sum_{f=1}^{F} \sum_{d=1}^{D} |(y_{pred})_{f,d} - (y_{tar})_{f,d}|,
\end{equation}
where $F$ denotes the number of frames and $D$ the dimensionality of the mel-frequency bins. During inference, an ODE solver generates mel-spectrograms from Gaussian noise, conditioned on the speaker embeddings and the final semantic representation $Q_{fin}$.
\subsection{Text-to-Emotion (T2E)}
We achieve the effect of natural language emotion control through the following steps.


First, we define seven basic emotions: $\mathcal{E}$ = \{\text{Anger}, \text{Happiness}, \text{Fear}, \text{Disgust}, \text{Sadness}, \text{Surprise}, \text{Neutral}\}. 
For each emotion $e_i \in \mathcal{E}$, we extract embeddings from several relevant emotional audio samples using the pre-trained emotion perceiver conditioner in the T2S, forming a fixed emotion embedding set $\mathcal{V}$.

Then, we use the large language model Deepseek-r1 as a teacher to map a text input $t$ to a 7-dimensional emotion probability distribution:
\begin{equation}
p = \text{Deepseek-r1}(t) \in \Delta^7,
\end{equation}
where $\Delta^7$ is the 7-dimensional probability simplex ($\sum_{i=1}^7 p_i = 1$, $p_i \geq 0$). To enable efficient inference with smaller models, we apply knowledge distillation to transfer the teacher's behavior to a smaller student model Qwen-3-1.7b.

We construct a training dataset of 1000 text-distribution pairs using two types of prompts with Deepseek-r1:
\begin{itemize}
  \item \textbf{Descriptive}: ``Please generate descriptive sentences that express \textit{\{emotion\}}.''
  \item \textbf{Script-like}: ``Please generate script-like utterances that express \textit{\{emotion\}}.''
\end{itemize}
For each generated sentence, we use a classification prompt to obtain the corresponding emotion distribution: \\
``Given the input sentence, return a JSON object with probabilities for each of the 7 emotions. Probabilities must sum to 1 and be rounded to two decimal places.''

Using this dataset, we fine-tune Qwen-3-1.7b via LoRA. The training objective is to minimize the cross-entropy loss between the student model's predictions and the teacher-provided distributions:
\begin{equation}
\min_{\phi} \mathbb{E}_{(t, p) \sim \mathcal{D}} \left[ \text{CrossEntropy}\left( \text{Qwen-3}_{\theta + \phi}(t), p \right) \right],
\end{equation}
where $\theta$ denotes the original parameters of Qwen-3-1.7b and $\phi$ represents the LoRA parameters. $t$ refers to input text samples from the dataset $\mathcal{D}$, while $p$ denotes the soft probability distributions generated by the teacher model. After training, the distilled Qwen-3-1.7b model can efficiently replace Deepseek-r1 during inference with significantly reduced computational cost.

Next step, the emotion vector $e_{\text{input}}$ is computed as a weighted average over the emotion embedding set $\mathcal{V}$:
\begin{equation}
e_{\text{input}} = \sum_{e \in \mathcal{E}} p_e \cdot \frac{1}{|\mathcal{V}_e|} \sum_{v \in \mathcal{V}_e} v.
\end{equation}

Finally, this emotion vector is fed as a prompt into the T2S model, enabling the generation of speech with the desired emotional characteristics.

\begin{table*}[ht]
\centering
\caption{Zero-Shot Performance Comparison of Various Systems on Different Datasets}
\label{tab:performance_comparison}
\begin{tabular}{@{}llccccc@{}}
\toprule
\textbf{Dataset} & \textbf{Model} & \textbf{SS$\uparrow$} & \textbf{WER(\%)$\downarrow$} & \textbf{SMOS$\uparrow$} & \textbf{PMOS$\uparrow$} & \textbf{QMOS$\uparrow$} \\
\midrule

\multirow{8}{*}{\begin{tabular}{@{}c@{}}\textbf{LibriSpeech} \\ \textbf{test-clean}\end{tabular}}
& Ground Truth & 0.833 & 3.405 & 4.02$_{\pm0.22}$ & 3.85$_{\pm0.26}$ & 4.23$_{\pm0.12}$ \\
& MaskGCT & 0.790 & 7.759 & 4.12$_{\pm0.09}$ & 3.98$_{\pm0.11}$ & 4.19$_{\pm0.19}$ \\
& F5-TTS & 0.821 & 8.044 & 4.08$_{\pm0.21}$ & 3.73$_{\pm0.27}$ & 4.12$_{\pm0.13}$ \\
& CosyVoice2 & 0.843 & 5.999 & 4.02$_{\pm0.22}$ & 4.04$_{\pm0.28}$ & 4.17$_{\pm0.25}$ \\
& SparkTTS & 0.756 & 8.843 & 4.06$_{\pm0.20}$ & 3.94$_{\pm0.21}$ & 4.15$_{\pm0.16}$ \\
& IndexTTS & 0.819 & 3.436 & 4.23$_{\pm0.14}$ & 4.02$_{\pm0.18}$ & 4.29$_{\pm0.22}$ \\
& IndexTTS2 & 0.870 & \textbf{3.115} & \textbf{4.44}$_{\pm0.12}$ & \textbf{4.12}$_{\pm0.17}$ & \textbf{4.29}$_{\pm0.14}$ \\
\cmidrule(lr){2-7} 
& - GPT latent & \textbf{0.887} & 3.334 & 4.33$_{\pm0.10}$ & 4.10$_{\pm0.12}$ & 4.17$_{\pm0.22}$ \\
\midrule

\multirow{8}{*}{\begin{tabular}{@{}c@{}}\textbf{SeedTTS} \\ \textbf{test-en}\end{tabular}}
& Ground Truth & 0.820 & 1.897 & 4.21$_{\pm0.19}$ & 4.06$_{\pm0.25}$ & 4.40$_{\pm0.15}$ \\
& MaskGCT & 0.824 & 2.530 & 4.35$_{\pm0.20}$ & 4.02$_{\pm0.24}$ & 4.50$_{\pm0.17}$ \\
& F5-TTS & 0.803 & 1.937 & 4.44$_{\pm0.14}$ & 4.06$_{\pm0.21}$ & 4.40$_{\pm0.12}$ \\
& CosyVoice2 & 0.794 & 3.277 & 4.42$_{\pm0.26}$ & 3.96$_{\pm0.24}$ & 4.52$_{\pm0.15}$ \\
& SparkTTS & 0.755 & 1.543 & 3.96$_{\pm0.23}$ & 4.12$_{\pm0.22}$ & 3.89$_{\pm0.20}$ \\
& IndexTTS & 0.808 & 1.844 & \textbf{4.67}$_{\pm0.16}$ & \textbf{4.52}$_{\pm0.14}$ & \textbf{4.67}$_{\pm0.19}$ \\
& IndexTTS2 & 0.860 & \textbf{1.521} & 4.42$_{\pm0.19}$ & 4.40$_{\pm0.13}$ & 4.48$_{\pm0.15}$ \\
\cmidrule(lr){2-7} 
& - GPT latent & \textbf{0.879} & 1.616 & 4.40$_{\pm0.22}$ & 4.31$_{\pm0.17}$ & 4.42$_{\pm0.20}$ \\
\midrule

\multirow{8}{*}{\begin{tabular}{@{}c@{}}\textbf{SeedTTS} \\ \textbf{test-zh}\end{tabular}}
& Ground Truth & 0.776 & 1.254 & 3.81$_{\pm0.24}$ & 4.04$_{\pm0.28}$ & 4.21$_{\pm0.26}$ \\
& MaskGCT & 0.807 & 2.447 & 3.94$_{\pm0.22}$ & 3.54$_{\pm0.26}$ & 4.15$_{\pm0.15}$ \\
& F5-TTS & 0.844 & 1.514 & 4.19$_{\pm0.21}$ & 3.88$_{\pm0.23}$ & 4.38$_{\pm0.16}$ \\
& CosyVoice2 & 0.846 & 1.451 & 4.12$_{\pm0.25}$ & 4.33$_{\pm0.19}$ & 4.31$_{\pm0.21}$ \\
& SparkTTS & 0.683 & 2.636 & 3.65$_{\pm0.26}$ & 4.10$_{\pm0.25}$ & 3.79$_{\pm0.18}$ \\
& IndexTTS & 0.781 & 1.097 & 4.10$_{\pm0.09}$ & 3.73$_{\pm0.23}$ & 4.33$_{\pm0.17}$ \\
& IndexTTS2 & 0.865 & \textbf{1.008} & 4.44$_{\pm0.17}$ & \textbf{4.46}$_{\pm0.11}$ & \textbf{4.54}$_{\pm0.08}$ \\
\cmidrule(lr){2-7} 
& - GPT latent & \textbf{0.890} & 1.261 & \textbf{4.44}$_{\pm0.13}$ & 4.33$_{\pm0.15}$ & 4.48$_{\pm0.17}$ \\
\midrule

\multirow{8}{*}{\begin{tabular}{@{}c@{}}\textbf{AIShell-1} \\ \textbf{test}\end{tabular}}
& Ground Truth & 0.847 & 1.840 & 4.27$_{\pm0.19}$ & 3.83$_{\pm0.25}$ & 4.42$_{\pm0.07}$ \\
& MaskGCT & 0.598 & 4.930 & 3.92$_{\pm0.03}$ & 2.67$_{\pm0.08}$ & 3.67$_{\pm0.07}$ \\
& F5-TTS & 0.831 & 3.671 & 4.17$_{\pm0.30}$ & 3.60$_{\pm0.25}$ & 4.25$_{\pm0.22}$ \\
& CosyVoice2 & 0.834 & 1.967 & 4.21$_{\pm0.23}$ & 4.33$_{\pm0.19}$ & 4.40$_{\pm0.21}$ \\
& SparkTTS & 0.593 & 1.743 & 3.48$_{\pm0.22}$ & 3.96$_{\pm0.16}$ & 3.79$_{\pm0.20}$ \\
& IndexTTS & 0.794 & \textbf{1.478} & 4.48$_{\pm0.18}$ & 4.25$_{\pm0.19}$ & 4.46$_{\pm0.07}$ \\
& IndexTTS2 & 0.843 & 1.516 & \textbf{4.54}$_{\pm0.11}$ & \textbf{4.42}$_{\pm0.17}$ & \textbf{4.52}$_{\pm0.17}$ \\
\cmidrule(lr){2-7} 
& - GPT latent & \textbf{0.868} & 1.791 & 4.33$_{\pm0.22}$ & 4.27$_{\pm0.26}$ & 4.40$_{\pm0.19}$ \\
\bottomrule
\end{tabular}
\end{table*}

\section{Experiments}
\subsection{Experimental Settings}

\begin{table*}[ht]
\centering
\caption{Performance Comparison of Various Systems on the Emotional Test Dataset}
\label{tab:emotion_comparison}
\begin{tabular}{l c c c c c c c}
\toprule
\textbf{Model} & \textbf{SS$\uparrow$} & \textbf{WER(\%)$\downarrow$} & \textbf{ES$\uparrow$} 
                & \textbf{SMOS$\uparrow$} & \textbf{EMOS$\uparrow$} 
                & \textbf{PMOS$\uparrow$} & \textbf{QMOS$\uparrow$} \\
\midrule
MaskGCT & 0.810 & 4.059 & 0.841 
        & 3.42$_{\pm0.36}$ & 3.37$_{\pm0.42}$ & 3.04$_{\pm0.40}$ & 3.39$_{\pm0.37}$ \\
F5-TTS & 0.773 & 3.053 & 0.757 
        & 3.37$_{\pm0.40}$ & 3.16$_{\pm0.32}$ & 3.13$_{\pm0.30}$ & 3.36$_{\pm0.29}$ \\
CosyVoice2 & 0.803 & 1.831 & 0.802 
           & 3.13$_{\pm0.32}$ & 3.09$_{\pm0.33}$ & 2.98$_{\pm0.35}$ & 3.28$_{\pm0.22}$ \\
SparkTTS & 0.673 & 2.299 & 0.832 
         & 3.01$_{\pm0.26}$ & 3.16$_{\pm0.24}$ & 3.21$_{\pm0.28}$ & 3.04$_{\pm0.18}$ \\
IndexTTS & 0.649 & \textbf{1.136} & 0.660 
         & 3.17$_{\pm0.39}$ & 2.74$_{\pm0.36}$ & 3.15$_{\pm0.36}$ & 3.56$_{\pm0.27}$ \\
IndexTTS2 & 0.836 & 1.883 & 0.887
          & \textbf{4.24$_{\pm0.19}$} & \textbf{4.22$_{\pm0.12}$} & \textbf{4.08$_{\pm0.20}$} & \textbf{4.18$_{\pm0.10}$} \\
\midrule
 - GPT latent & \textbf{0.869} & 2.766 & \textbf{0.888} 
          & 4.15$_{\pm0.20}$ & 4.15$_{\pm0.19}$ & 4.02$_{\pm0.20}$ & 4.03$_{\pm0.11}$ \\
 - Training strategy & 0.773 & 1.362 & 0.689 
          & 3.44$_{\pm0.29}$ & 2.82$_{\pm0.35}$ & 3.83$_{\pm0.33}$ & 3.69$_{\pm0.18}$ \\
\bottomrule
\end{tabular}
\end{table*}

\begin{table*}[ht]
\centering
\caption{Comparison of Natural Language-Based Emotion Control with CosyVoice2}
\label{tab:natural_language_comparison}
\begin{tabular}{l c c c c}
\toprule
\textbf{Model} & \textbf{SMOS$\uparrow$} & \textbf{EMOS$\uparrow$} 
                & \textbf{PMOS$\uparrow$} & \textbf{QMOS$\uparrow$} \\
\midrule
CosyVoice2  & 2.973$_{\pm0.26}$ & 3.339$_{\pm0.30}$ & 3.679$_{\pm0.19}$ & 3.429$_{\pm0.24}$ \\
IndexTTS2  & \textbf{3.875$_{\pm0.21}$} & \textbf{3.786$_{\pm0.24}$} & \textbf{4.143$_{\pm0.13}$} & \textbf{4.071$_{\pm0.15}$} \\
\bottomrule
\end{tabular}
\end{table*}

\subsubsection{Datasets:} We trained our model using 55K hours of data, including 30K Chinese data and 25K English data. Most of the data comes from Emilia dataset~\cite{he2024emilia}, in addition to some audiobooks and purchasing data. A total of 135 hours of emotional data came from 361 speakers, of which 29 hours came from the ESD dataset~\cite{esddataset} and the rest from commercial purchases. To validate the fundamental capabilities of TTS systems, we evaluated our model on four benchmarks: (1) SeedTTS test-en ~\cite{anastassiou2024seed}, introduced in SeedTTS containing 1,000 utterances from the Common Voice dataset; (2) SeedTTS test-zh ~\cite{anastassiou2024seed}, 2,000 utterances sourced from DiDiSpeech~\cite{guo2021didispeech}; (3) LibriSpeech-test-clean \cite{panayotov2015librispeech}, 2,620 randomly selected utterances from the LibriSpeech corpus; and (4) AISHELL-1 \cite{bu2017aishell}, 1,000 utterances randomly sampled from the AISHELL-1 dataset. To better assess emotional modeling capability, we recruited 12 speakers (5 males and 7 females) to record an emotional test set. Each speaker recorded 3 sentences for each of the 7 emotional categories.

\subsubsection{Evaluation Metrics:} Objectively, speech intelligibility is evaluated using word error rate (WER), with FunASR \cite{funasr} for Chinese content and Whisper \cite{whisper} for English. Speaker similarity (SS) is computed as the cosine similarity between speaker embeddings from FunASR’s pretrained speaker recognition model, while emotion similarity (ES) is calculated using emotion representations from the open-source emotion2vec \cite{emotion2vec} model. Subjective evaluation is conducted through a multi-dimensional Mean Opinion Score (MOS) framework, where Similarity MOS (SMOS), Prosody MOS (PMOS), Quality MOS (QMOS), and Emotion MOS (EMOS) assess speaker similarity, prosody, audio quality, and emotional fidelity respectively, each rated on a 1–5 scale.




\subsubsection{Baseline:} We compared our model with state-of-the-art zeroshot TTS systems, including MaskGCT \cite{wang2024maskgct}, F5-TTS \cite{chen2024f5}, CosyVocie2 \cite{du2024cosyvoice2}, SparkTTS \cite{wang2025spark} and the original IndexTTS \cite{deng2025indextts} model. In addition, we conduct two ablation experiments to validate the architectural design and training methodology of IndexTTS2:
\textbf{(1) GPT latent enhancement removal. }
This experiment ablates the GPT-derived latent feature enhancement to evaluate its functional contribution in the S2M module.
\textbf{(2) Training strategy ablation. }
This experiment ablates additional training strategies to evaluate its contribution to highly expressive emotional speech synthesis.

\subsubsection{Training Hyperparameter Details:} We trained IndexTTS2 on 8 NVIDIA A100 80GB GPUs using the AdamW optimizer with an initial learning rate of 2e-4. Our model was trained for a total of three weeks. We used the same text tokenizer as IndexTTS and adopted the semantic codec from the MaskGCT model.

\subsection{Experiment Results}


\subsubsection{Basic Competence Comparison:} 
We evaluated IndexTTS2 on standard test sets (LibriSpeech-test-clean, SeedTTS test-zh/en, and AIShell-1 test)\footnote{To ensure cross-dataset evaluation consistency, we re-implemented some published experiments. Observed minor performance variations—attributable to inherent model fluctuations or using FunASR-provided speaker feature replacements—remain within acceptable ranges, preserve overall rankings, and validate original results.}. As shown in Table \ref{tab:performance_comparison}, compared to five representative models (MaskGCT, F5-TTS, CosyVoice2, SparkTTS, and the original IndexTTS),  IndexTTS2 achieves SOTA performance in objective evaluation across most test sets, with only marginal underperformance on AIShell-1 relative to the ground truth and IndexTTS. In subjective evaluation, IndexTTS2 outperforms all baseline models except for a slight underperformance against IndexTTS on SeedTTS test-en. Results from the ablation experiment show that removing the GPT latent enhancement consistently improves SS while degrading WER across datasets, and the ablated model receives lower subjective scores across the board compared to IndexTTS2. Notably, the subjective speaker similarity MOS (SMOS) indicate that despite the slight drop in SS, the enhanced model is perceived by human listeners as more similar to the target speaker. These findings confirm the importance of GPT latents in enhancing semantic clarity.

\subsubsection{Emotional Performance Comparison:} 

We evaluated IndexTTS2's emotional expressiveness on our constructed emotional dataset using relevant metrics. As shown in Table \ref{tab:emotion_comparison}, IndexTTS2 achieves the highest scores across all four subjective evaluation dimensions, demonstrating superior emotional rendering capabilities. Examining the objective metrics, compared to five baseline models, IndexTTS2 shows leading performance in SS and ES, except for higher WER than IndexTTS. In the ablation setting, while IndexTTS2 exhibits slightly lower SS and ES than the variant without GPT latent enhancement, the gap in SS is minimal and the difference in ES (0.001) is practically insignificant; however, it maintains a clear advantage in WER and achieves superior performance across all subjective metrics. This indicates that the GPT latent enhancement in the S2M module play a crucial role in maintaining speech clarity and articulation under high emotional expressiveness. In contrast, removing the three-stage training strategy severely degrades emotional expressiveness, resulting in substantial performance drops across all metrics except WER. Overall, these results demonstrate that IndexTTS2, with its multi-stage training incorporating GRL-based emotion disentanglement and GPT fusion, effectively balances emotional expressiveness with speech clarity, achieving state-of-the-art performance in emotional speech synthesis while maintaining exceptional textual accuracy.



\subsubsection{Evaluation of Natural Language-Controlled Emotional Synthesis:} 

We evaluated the T2E module's effectiveness for natural language emotion control using a constructed test set (Table \ref{tab:natural_language_comparison}). The test set included texts with half manually assigned emotion prompts and half using target texts as prompts. Through double-blind human evaluation across four metrics (timbre similarity, emotion similarity, rhythm and audio quality), our approach outperformed CosyVoice2 in all aspects, demonstrating superior natural language-based emotion control capabilities. This confirms its enhanced ability to align speech synthesis with specified emotional contexts while maintaining consistent performance.

\begin{table}[ht]
    \centering
    \caption{Token Number Error Rate for Duration Control with Different Settings(\%)}
    \label{tab:token_control} 
    \begin{tabular}{l c c c c c}
        \toprule
        \textbf{Dataset} & \textbf{*1} & \textbf{*0.75} & \textbf{*0.875} & \textbf{*1.125} & \textbf{*1.25} \\
        \midrule
        \multirow{2}{*}{\makecell{\textbf{SeedTTS} \\ \textbf{test-zh}}} 
            & \multirow{2}{*}{0.019} & \multirow{2}{*}{0.067} & \multirow{2}{*}{0.023} & \multirow{2}{*}{0.014} & \multirow{2}{*}{0.018} \\
        & & & & & \\
        \cmidrule{1-6}
        \multirow{2}{*}{\makecell{\textbf{SeedTTS} \\ \textbf{test-en}}} 
            & \multirow{2}{*}{0.015} & \multirow{2}{*}{0} & \multirow{2}{*}{0.009} & \multirow{2}{*}{0.023} & \multirow{2}{*}{0.013} \\
        & & & & & \\
        \bottomrule
    \end{tabular}
\end{table}

\begin{table}[htbp]
\centering
\caption{MOS Scores for Different Models under Duration Control}
\begin{tabular}{@{}p{0.08\textwidth}p{0.08\textwidth}p{0.07\textwidth}p{0.07\textwidth}p{0.07\textwidth}@{}}
\toprule
\textbf{Datasets} & \textbf{Model} & \textbf{SMOS$\uparrow$} & \textbf{PMOS$\uparrow$} & \textbf{QMOS$\uparrow$} \\
\midrule

\multirow{5}{*}{\begin{tabular}{@{}c@{}}\textbf{SeedTTS} \\ \textbf{test-zh}\end{tabular}}
& GT             & 3.82$_{\pm0.23}$ & 3.72$_{\pm0.19}$ & 3.96$_{\pm0.06}$  \\
& MaskGCT        & 4.04$_{\pm0.18}$ & 4.16$_{\pm0.06}$ & 3.66$_{\pm0.11}$ \\
& F5-TTS             & 4.32$_{\pm0.15}$ & 4.04$_{\pm0.15}$ & 4.32$_{\pm0.16}$ \\
& IndexTTS2      & \textbf{4.56}$_{\pm0.08}$ & \textbf{4.38}$_{\pm0.12}$ & \textbf{4.42}$_{\pm0.02}$ \\

\midrule

\multirow{5}{*}{\begin{tabular}{@{}c@{}}\textbf{SeedTTS} \\ \textbf{test-en}\end{tabular}}
& GT             & 4.32$_{\pm0.26}$ & 4.34$_{\pm0.05}$ & 4.42$_{\pm0.11}$ \\
& MaskGCT        & \textbf{4.54$_{\pm0.16}$} & 4.24$_{\pm0.08}$ & 4.44$_{\pm0.13}$ \\
& F5-TTS             & 4.34$_{\pm0.18}$ & 4.24$_{\pm0.06}$ & 4.26$_{\pm0.09}$ \\
& IndexTTS2      & 4.48$_{\pm0.09}$ & \textbf{4.46}$_{\pm0.18}$ & \textbf{4.44$_{\pm0.05}$} \\

\bottomrule
\end{tabular}
\label{tab:mos_scores}
\end{table}

\subsubsection{Duration-Specified Speech Synthesis Evaluation:}


We evaluated IndexTTS2's duration control accuracy on SeedTTS test-zh and test-en using five experimental setups with duration scalings (original, 0.75×, 0.875×, 1.125×, and 1.25×). Results in Table \ref{tab:token_control} show minimal token number error rates (\textless 0.02\% for original durations and \textless 0.03\% for 0.875×/1.125×), with only a slight increase to 0.067\% on SeedTTS test-zh for larger scaling factors (0.75×). These findings indicate an almost negligible gap between generated tokens and target durations, demonstrating IndexTTS2's precise control over speech synthesis timing.  

\begin{figure}[tbp] 
    \centering 
    \begin{subfigure}{0.23\textwidth} 
        \centering 
        \includegraphics[width=\linewidth]{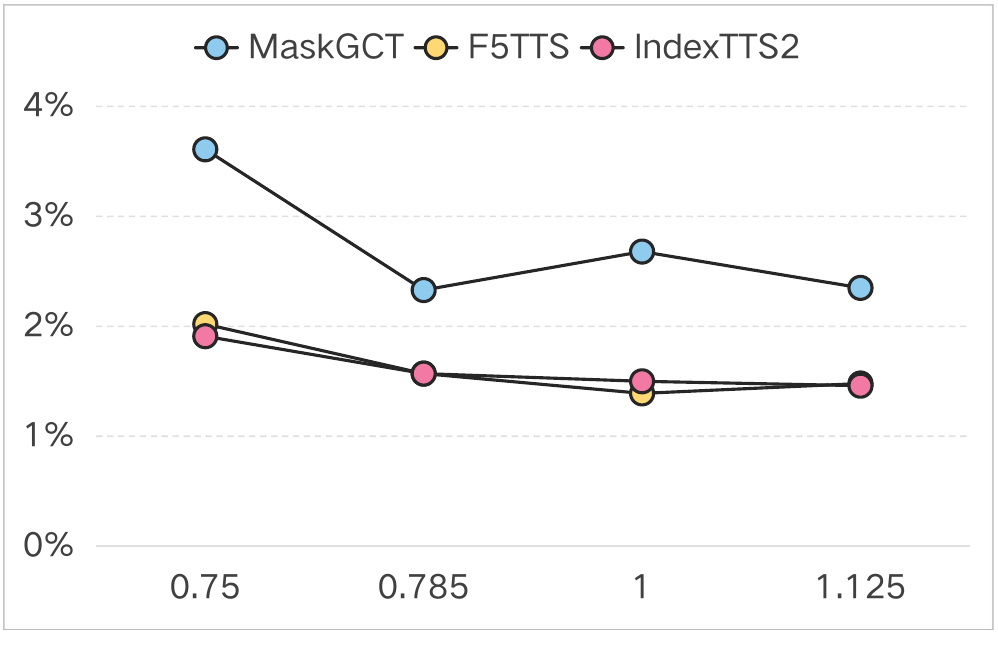} 
        \caption{SeedTTS test-en} 
        \label{fig:seeden-cropped} 
    \end{subfigure}
    \hfill 
    \begin{subfigure}{0.23\textwidth} 
        \centering
        \includegraphics[width=\linewidth]{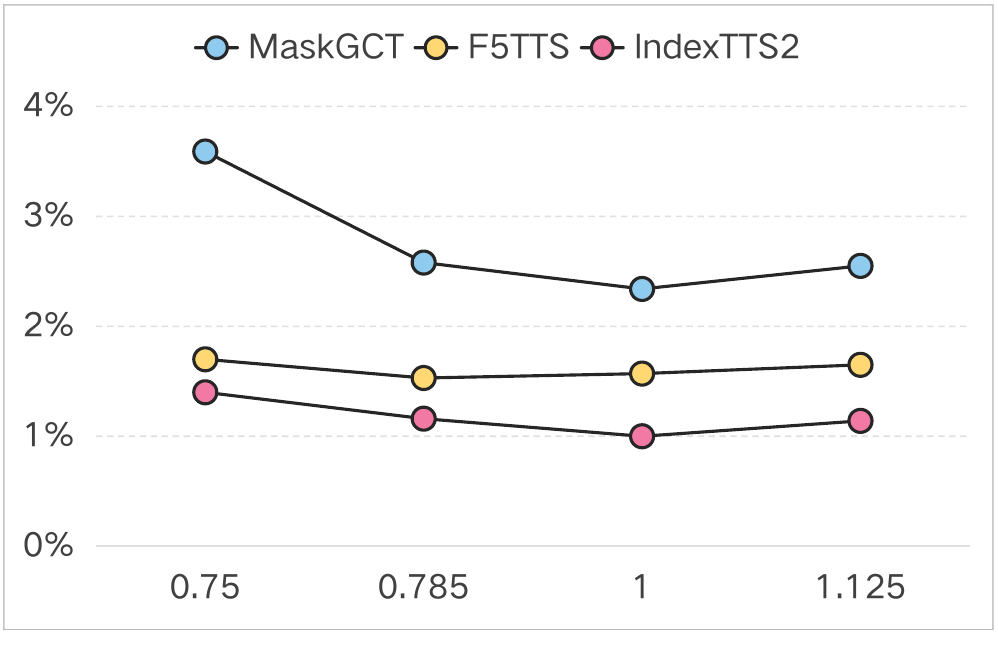} 
        \caption{SeedTTS test-zh} 
        \label{fig:seedzh-cropped} 
    \end{subfigure}
    \caption{Comparison of WER for duration control section.} 
    \label{fig:combined_seeds} 
\end{figure}

We further assessed speech quality under duration control by comparing WER. As shown in Figure~\ref{fig:combined_seeds}, IndexTTS2 matches F5-TTS on test-en and significantly outperforms MaskGCT. On test-zh, IndexTTS2 surpasses F5-TTS by ~0.5 pp and MaskGCT by ~2 pp, with only a marginal drop in performance under scaled durations. To investigate the prosodic advantages of autoregressive modeling under fixed duration control, we conducted a comparison between IndexTTS2 and non-autoregressive TTS systems using SMOS, PMOS, and QMOS metrics. Results in Table \ref{tab:mos_scores} show that IndexTTS2 achieves superior prosody and speech quality.

\section{Conclusion}

In this work, we present IndexTTS2, a zero-shot speech synthesis system that enhances duration modeling, emotional expressiveness, and phonetic clarity via an innovative autoregressive architecture with optimized training. Featuring a unique duration control for precise timing and a mechanism to decouple emotional and speaker features, IndexTTS2 facilitates emotion-specific speech generation from reference audio. An LLM-driven module matches emotion vectors for language-based modulation, ensuring natural expression. Combined with specialized training and data augmentation strategies, the model achieves SOTA-level performance in high-expressive emotional restoration. Efficient in zero-shot settings, IndexTTS2 produces expressive speech with controllable duration and emotions, advancing voice solutions for animated dubbing and video narration while pushing the boundaries of speech synthesis technology.

\bibliography{aaai2026}


\end{document}